# A Path to Loving

John BEVERLEY[a,b,c] and Regina HURLEY[a,b]

[a] *National Center for Ontological Research*
[b] *University at Buffalo*
[c] *Institute for Artificial Intelligence and Data Science*

**Abstract.** This work lays the foundations for a rigorous ontological characterization for love, addressing its philosophical complexity and scientific relevance - with particular emphasis on psychology and sociology – as well as highlighting ways in which such characterization enhances relevant AI-based applications. The position defended here is that love is best understood as a concatenation of passive sensations (e.g., emotional arousal) and active evaluative judgments (e.g., perceiving the beloved as valuable), in the interest of balancing the involuntary aspects of love with its rational accountability. To provide a structured foundation, the paper draws on Basic Formal Ontology (BFO) and other applied ontological methods to differentiate various senses of love. This work engages with objections to the love understood as concatenation, particularly concerning the relationship between sensation and judgment. A causal correlation model is defended, ensuring that the affective and cognitive components are linked. By offering a precise and scalable ontological account, this work lays the foundation for future interdisciplinary applications, making love a subject of formal inquiry in ontology engineering, artificial intelligence, and among the sciences.

## 1. Introduction

Love is core to the lived experiences of most, if not all; it is pervasive and intimate, beautiful and tragic. As one might expect, it is a popular research area [1]. Life science researchers, for example, explore neurobiological underpinnings of love [2, 3, 4], the endocrinological underpinnings of romantic love [5, 6], and the evolutionary impact on fitness [7, 8, 9], among many other topics. Psychologists investigate the impact of loving on the accuracy of judgments in relationships [10], pathological facets of so-called "love addiction" [11, 12], and loving kindness meditation interventions [13]. Love is a topic of interest across numerous other disciplines, such as sociology [14], ecology [15], and philosophy [16, 17]. Given its pervasiveness both in our experiences and as a topic of scientific investigation, it is a wonder that love has not been the target of any robust ontological analysis. Where one finds mention of love in ontology engineering literature it is almost always as an example in passing [18, 19, 20, 21].[1] Perhaps this owes to the almost overwhelming manifestations of the phenomena, e.g. parental love, sibling love, love between friends, erotic love, etc., suggesting the best analysis one can hope for ultimately ends up as family resemblance [23].[2] Perhaps this is so. In any event, the realm of romantic love is broad enough to cover much of the research cited, and structural features of it gleaned from natural language suggest susceptibility to ontological analysis.

---

[1] Though see footnote 12 in [22] for a bit more.
[2] Or perhaps a *focal analysis* of the sort Aristotle employs in discussion of friendship.

"Love" is, of course, spoken of in many ways. Sally may be a lover, loved, loveable, lovely, loving, or all the above. Similarly, John and Sally may be lovers, in love, loving, etc. These cursory ordinary speech examples suggest we speak of love ambiguously. More specifically, they suggest we speak of love and love-related notions sometimes as if they are properties and other times as if they are relational. For example, saying Sally is lovable, lovely, etc., suggests reference to properties of Sally. Alternatively, Sally as a lover, loved, loving, or in love with John, etc., suggests a relation between Sally and some other entity. There is also a process sense of love that one finds reference to when, say, claiming that Sally is in a (continued) state of loving or (being) in love. Natural language being what it is, these remarks do not exhaust all the ways in which "love" and its variants may be used but do provide a starting point for discussion.[3]

In what follows, we take up the laborious first steps towards an ontological foundation of love. We first provide hallmarks of love and loving, fending off objections from philosophical and conceptual analysis. We ultimately defend an account of love as temporally extended and grounded in human agency, involving both evaluative judgments and sensations. Though we occupy ourselves largely with contemporary philosophical research on love, our goal is ultimately practical; as such, we close by setting up ontological scaffolding and invite the broader community to collaborate on developing robust ontological characterizations of love as they appear across disciplines, applications, databases, and our lives.

## 2. Hallmarks of Love

In this section, we outline hallmarks of love as preamble to the ontological analysis of the foundations of loving to follow.

### 2.1. Love as Active and Passive

Love has passive and active aspects [25, 26]. Loving seems *passive* in the sense that it is not under our immediate control. We may find ourselves in love but seem unable to, say, reason our way to it. On the other hand, we seem answerable for loving in the sense that one might felicitously request justification for loving [26, 27]. In this respect, love seems *active*. This is not to say one is expected to be able to provide exhaustive reasons for why they love someone, as an iron-clad deductive argument. We are rarely able to provide such strong justification for anything; it is certainly not required here. The idea is rather that when one finds oneself loving, one is open to requests for something more than an explanation for how it happened, but indeed an explanation for why. Suppose Sally loves John because she does not understand what she values, she is unreflective about such things, and John was regularly around and generally supportive of her. Presumably, Sally will lack justification for loving John even if she is answerable for loving. Suppose Sally loves John, but his values change significantly over the course of their relationship, so he is nearly unrecognizable. We might ask whether Sally should continue to love John; Sally is in this sense answerable. Where one comes down on what Sally should do in such cases is not our topic. These remarks simply illustrate that Sally is answerable for her love, insofar as it is felicitous to request justification. In this sense, love seems active.

---

[3] Natural language intuitions should not decide ontology content, but if an ontology representation deviates from them, an explanation is owed [24].

These aspects of love reveal a puzzle: How can Sally be answerable for loving if it is passive? Consider a potentially parallel passive phenomenon: perceiving. Sally is arguably not answerable for perceiving an apple. That is, one could not felicitously ask for Sally's *justification* for perceiving an apple. If asked, Sally would be correct to simply respond "because I do" or perhaps "because I happened to be looking that way." To be fair, Sally might similarly respond to the request for justification for loving but such responses seem inadequate here. This is because the question posed with respect to Sally loving plausibly targets why Sally loves John. For Sally to parallel justification for perceiving, her answer should refer to both how and why she found herself in the present situation. Such an answer will undoubtably be rather complex, not the least bit because loving takes time, in contrast to perception which is immediate.

It is here we find a path to address the puzzle of the passiveness and activeness of love. One often has plenty of opportunities to inhibit loving, which suggests if one finds oneself loving, one at least did not prevent a so-called "fall" into it. If this is correct, then the felicity of asking for justification with respect to love is, more specifically, a request for justification for why one allowed oneself to fall in love [27]. Put another way, we appear to have *inhibitory control* over loving. Here we find the grounds for love's active aspect.

This sort of interplay between inhibitory control and justification is not unique to loving. Generally, when agents find themselves in certain passive bodily states and answerable for being so, the agent has or had some manner of inhibitory control. Suppose that John intends to lose weight, and while adopting a new diet that finds him eating less than usual, expresses to his trainer Sally that he now regularly feels hungry. Sally might ask for justification for John's hunger. A charitable reading of Sally's request – along the lines of the charitable reading one takes when requesting justification for love – has her request aimed at why John did not previously take steps to inhibit his appetite. For had John dieted previously, he would not be so hungry, and so in that sense his present state of hunger was under his inhibitory control. In other words, despite the passiveness of being hungry, one may be answerable for being in that state, based on one's prior inhibitory control. It is in this sense that even hunger can be understood as active.

*2.2. Concatenation View*

A natural explanation for the preceding observations holds that loving is the concatenation of attitudes, one active and one passive. The passive component is a strong positive sensation, something like a fluttering in one's stomach. The active component is a judgment about the individual loved, say, that they are valuable.[4] More specifically, the "Concatenation View" (CV) maintains:

> **CV** For agents *S*, *P* and temporal instant *t*, '*S* loves *P*' is true at *t* just in case:
>     **(i)**  *S* has strong positive sensations about *P* at *t*
>     **(ii)** *S* judges *P* is valuable at *t*

---

[4] [25] advocates a version of this view, calling the non-cognitive aspect sub-rational and the cognitive aspect – which is open to justificatory challenge – rational.

This, at least, is how **CV** is typically presented [25, 27]. With respect to **(i)**, the claim is that, say, Sally has a strong positive sensation about John at a time. As we will see below, however, characterizing Sally's strong positive sensation as "about" John is misleading, since strictly speaking sensations – unlike, say, perceptions – are not "about" anything, as they lack content.[5] We should thus understand "about" in a looser sense in **CV**, though a sense needing to be explicated.

One may also worry that commitment to **(i)** makes the feelings associated with love analogous to headaches [27]. More generally, advocates of **CV** owe an explanation for how to distinguish sensations claimed in **(i)** from other sensations. The concern seems to stem from the felicity of saying being in love is to be pursued or avoided while this seems infelicitous when speaking about headaches or similar sensations. It is worth noting in response, that we need not be committed to the claim that all sensations are evaluable in the same way. I might evaluate my being hungry negatively if I am on a diet and feel shame about prior poor eating habits. With hunger as with love, but not so much with headaches, we have inhibitory control and so may felicitously evaluate corresponding sensations. Even though the sensations in **(i)** are analogous to headaches insofar as they are sensations, it does not follow that because headaches cannot be evaluated, then other sensations cannot too. We thus set this worry aside.

Turning to **(ii)**, the claim is that, say, Sally judges John is valuable at a time. Presumably, there are many ways in which one might judge another valuable, i.e. many determinates of value under a determinable. Putting aside judging valuable relative to some goal, we might distinguish judging someone valuable simply because they are an autonomous agent deserving respect, from judging someone valuable because – in addition – they are, for lack of a better description, special [28, 29]. We suspect advocates of **CV** would find satisfying **(ii)** too easy if agents need only value agents because autonomous and deserving respect. That is a judgment many would have about any agent. If that were the way to understand **(ii)**, then the only thing differentiating loving someone from valuing them as an autonomous agent would be whether **(i)** is satisfied or not. Perhaps this is defensible. But in the interest of being charitable, we interpret **CV** as involving both a certain species of positive sensation "about" an agent and a special determinate of the determinable value.

*2.3. Temporal Considerations*

Note the temporal restriction to instants in **CV**. According to **CV**, any time **(i)** and **(ii)** are satisfied, a loving relation exists. But this would make it far too easy, it seems, to love.[6] With that in mind, it seems charitable to understand **CV** in a slightly different manner, as a process involving at least two temporal instants as proper parts of the temporal interval over which it exists:

**CV2** For agents *S*, *P* and temporal interval *i*, '*S* loves *P*' is true at *i* just in case:
  **(i)**  *S* has strong positive sensations about *P* at *i*
  **(ii)** *S* judges *P* is valuable at *i*

---

[5] Content need not be propositional. "Content" is the genus, "propositional content" a species. Indeed, our paradigmatic cases involve *people* as content.
[6] This concern was raised by Kyla Ebels-Duggan in personal correspondence.

**CV2** as described presumes loving occurs at temporal intervals. One may wonder where this leaves states of loving at a time. But it is straightforward to view states of loving at a time as derivative from loving as a process over an interval. Temporal intervals have proper temporal point parts; loving over intervals has proper parts, some of which involve, we might say, loving states. Moreover, there seems to be correspondence between loving over an interval and the interval at which two individuals love. Similarly, there seems correspondence between loving states and temporal points at which two individuals love.

Unfortunately, **CV2** does not quite capture loving in the manner one might hope. For one may worry about the relationship between loving at an interval and at a point. First, it may be asked whether it is necessary according to **CV2** that S loves P *throughout* each of the temporal points comprising the relevant temporal interval. That is too restrictive as it would rule out many genuine cases of loving. Second, if loving throughout the interval is not necessary, then one might wonder whether it is sufficient according to **CV2** for *S* to count as loving *P* over an interval if there is one temporal point in that interval on which **(i)** and **(ii)** are satisfied. That is too permissive and would rule *in* many *im*plausible cases of loving.

What seems needed is a middle ground between these extremes. Proposing any specific, global, threshold - fixing on a number of loving states as needed to count one as loving according to **CV2** - would be problematically arbitrary. Fortunately, we need not venture into this territory. For our purposes, it suffices to assert there must be a strong positive correlation between loving over an interval and is love state temporal parts. More formally, we first define loving events:

> **(EVT)** Any temporal interval *i* and proper temporal part *t* of *i*, and for any agents *S*, *P* such that '*S* loves *P*' is true at *i* and **(i)** and **(ii)** are satisfied at *t* is a *love event*

In words, **(EVT)** defines a love event given '*S* loves *P*' is true as any proper state part of the loving process where **(i)** and **(ii)** are satisfied. This does not entail at every proper state part of the loving process that these conditions are satisfied. We define too the sum of loving events over an interval:

> **(SUM)** The disjoint mereological sum of love events over a given interval *i* is the *loving sum*

We also need:

> **(CPL)** The complement of the loving sum over a given interval *i* is the *loving complement*

Finally, we use the preceding definitions to capture the needed positive correlation between loving over an interval and loving at a proper state part of that interval, since this correlation builds in the condition **(i)** and **(ii)** from **CV2**:

> **CV3** For agents *S*, *P* and temporal interval *i*, loving sum *s* and loving complement *c* of *i*, '*S* loves *P*' is true at *i* just in case: there is some real value *T* such that: $T < s/c$

Importantly, **CV3** allows that *S* might not always satisfy **(i)** and **(ii)**, yet still count as loving, and similarly is compatible with ruling out the possibility that *S* loves someone at only one instant, and therefore loves them over a larger interval.[7] *T* is underspecified by design, since offering specific values for when there is a sufficient number of loving states satisfying **(i)** and **(ii)** to count as a loving process would be too restrictive given how widely love can be understood. Rather, we rest here by observing there is some positive ratio within an interval at which '*S* loves *P*' is true such that there is more satisfaction of **(i)** and **(ii)** than not as understood by whatever value *T* takes.

*2.4. Causality Considerations*

Timing issues are not the only aspects worth clarifying. Note, **CV3** maintains conditions **(i)** and **(ii)** from **CV**, the first involving a mere positive sensation and the second involving a judgment of value. We observed "about" is likely best understood in a loose sense but also observed this should be explicated in more detail. We can sharpen the motivation for explication by observing since sensations do not have content – for our purposes, something they are *about* - a question arises over how exactly **(i)** and **(ii)** are related. For if sensations lack content but judgments have content, then what makes it the case that the positive sensations in **(i)** are "about" the same individual involved in **(ii)**? *If* sensations *did* have content, this would be no problem, since advocates of **CV3** could simply say one counts as loving in this sense when the contents are the same. This is not a viable option.

    A natural, but unhelpful, response is to claim **(i)** and **(ii)** are simply regularities involving the same individual.[8] However, this does little to address the objection and might lead to accidental cases of loving where individuals just so happen to satisfy **(i)** and **(ii)** as a matter of coincidence. What is needed is more than a mere regularity between **(i)** and **(ii)**. To address this, we revise the implicit condition **(i)** in **CV3**, changing "about" to "causal correlation" linking individuals to the strong positive sensations involved in this condition. Of course, this alone does not do much to bridge the gap between **(i)** and **(ii)**, since causal correlations may amount to regularities too; we still must bridge the content of **(ii)** to the causally correlated individuals in **(i)**.

    But advocates of **CV3** may make use of the independently defensible distinction between experiencing as an attitude one takes towards content and awareness or perception of that content [30]. This distinction can be motivated with examples like the following: *S* looks for cufflinks in a drawer but does not find them. *S* later realizes the cufflinks were in the drawer, having reflected on memories of the search, but for whatever reason *S* did not notice them during the search. Again, the plausibility of this scenario suggests a distinction between *experiencing content* and *being aware or perceiving that content*. Put another way, *S* experienced the cufflinks but was not aware of or did not perceive them. The distinction between experiencing and awareness can be motivated as explaining other phenomena as well. Suppose *S* has a headache, then begins a distracting activity during which *S* does not feel any pain. After the activity has concluded, however, *S* experiences a headache. An explanation for this scenario is that *S experiences* the same headache throughout but was *not aware* of the headache while engaged in the distracting activity.

---

[7]We are not concerned with putative counterexamples such as: Sally satisfies **(i)** and **(ii)** with respect to John at instant t, but then they both go out of existence; this should count as loving but does not according to **CV** since temporal intervals must have more than one proper part. We bite the bullet.

[8][27] considers this response on behalf of [25] but does not develop the response in any depth.

For our purposes, the distinction between experiencing and awareness can be used to link relevant sensations in **(i)** to objects of judgment in **(ii)** as follows:

**(1)** Sally experiences positive sensations causally correlated with John
**(2)** Sally judges the positive sensations of **(1)** are valuable
**(3)** Sally judges experiences of positive sensations causally correlated with John are valuable
**(4)** Sally judges John is valuable

Walking through the above link, we see in **(1)** Sally experiences positive sensations causally correlated with John that she then judges are valuable in **(2)**. The step from **(2)** to **(3)** has Sally judge the experiences in which the positive sensations causally correlated with John are found, as also valuable. Hence, in **(3)** Sally links the positive sensations to the content of an experience with John through judgment. Note too, **(3)** requires not only the experience of some content, but awareness of that content. This stems from commitment to the general claim that: *if S judges p, then S is aware of p*. With both valued positive sensations and John causally correlated with them in the content of Sally's experience, Sally then judges the cause of these positive sensations is valuable. Even more concisely, the move from **(1)** to **(4)** amounts to Sally experiencing positive sensations, judging they are valuable, recognizing they are involved as constituents in the content of a valuable experience, then judging an important causal constituent of those sensations is also valuable. Linking this back to **CV3**, advocates should say **(i)** and **(ii)** are satisfied when **(1)** - **(4)** are satisfied. With this, they have a clear connection between the loved individuals and sensations causally correlated with them, and so a clear bridge between **(i)** and **(ii)**.[9]

A final objection[10] one might offer against **CV3** is that it seems compatible with counting the following scenario as one in which Sally loves John: Suppose Sally ingests narcotics causing her to feel positive sensations causally correlated with John enough times to satisfy the ratio in **CV3**. Suppose each time Sally also judges that John is valuable in the relevant sense of **(ii)**. But suppose Sally only knows of him by description. Since Sally satisfies the condition of **CV3**, it follows Sally loves John over the relevant interval. But this is absurd, since Sally has never met John. As stated, we agree **CV3** is compatible with this possibility. It is, however, open to advocates of **CV3** to simply refine what it means to satisfy **(ii)** in an independently plausible manner, namely, by claiming that judgment of John's value in **(ii)** requires more than mere knowledge by description. Sally judging that a type of individual is valuable does not entail Sally judges a given token of that type is valuable; Sally judging that under a definite description is valuable does not entail Sally judges John referred to without knowledge by acquaintance is valuable as well. We observe this seems analogous to what one finds more generally in discussion of taste predicates [31]. It is infelicitous to claim a dish is delicious if one has never tasted it before. Taste predicates carry with them what is called an *acquaintance inference*, i.e. that the speaker has direct acquaintance with the item tasted. Advocates of **(ii)** might – and I think should – similarly assert the sense of value in **(ii)** also involves an acquaintance inference [32]. Since Sally in the above scenario has never met John, her putative judgments about his value do not satisfy **(ii)**, and hence she does not count as loving John according to **CV3**.

---

[9]One might object this requires too much of lovers, since to satisfy **(i)** and **(ii)** they must make value judgments about the beloved, and – when coupling this with the ratio condition of **CV2**, requires they perhaps make more such value judgments than not. The force of this worry trades on the specification of T, but advocates of **CV2** are not obviously committed to a problematic specification.

[10]This objection too was raised by [27] against [25].

*2.5. Summary*

The preceding discussion leaves us with a final characterization of love that will provide a foundation for our ontological characterization to follow.

> **CV4** For agents *S*, *P* and temporal interval *i*, loving sum *s* and loving complement *c* of *i*:
> **a.** '*S* loves *P*' is true at *i* just in case: there is some real value *T* such that: $T < s/c$,
> **b.** *s* consists of instances in which *S* experiences positive sensations causally correlated with *P* and judges these sensations and their associated experience of *P* as valuable,
> **c.** *S* has direct acquaintance with *P* that informs the *S*'s evaluative judgments that satisfy **b**

This formula clarifies the relationship between sensation and judgment by requiring structured causal correlations and acquaintance-based valuation. To establish an ontological foundation for further analysis, we anchor this characterization within the Basic Formal Ontology.

**3. Love in the Basic Formal Ontology**

Basic Formal Ontology (BFO) is an ISO/IEC 21838-2 top-level ontology [33] developed initially to facilitate consistent representations of data across various life science disciplines, though presently major users of BFO include developers in the Open Biological and Biomedical Ontologies (OBO) Foundry [34], the Industrial Ontologies Foundry (IOF) [35], and the Common Core Ontologies (CCO) suite of defense and intelligence ontologies [36]. BFO is used in over roughly 700 open-source ontology initiatives, where it provides a domain-neutral starting point for ontology development. Among these include ontologies scoped to psychology, behavioral science, sociology, neurobiology, and other areas relevant to scientific research on love. Accordingly, BFO provides an excellent starting point for ontologically representing our target phenomenon.

*3.1. Basic Formal Ontology Machinery*

Our task here is to show how **CV4** can be formally described using terms and relations from BFO. BFO terms represent highly general classes, the highest division being between **occurrent**[11] and **continuant**. **Occurrents** are entities that unfold over time and have temporal parts. They include **processes** which extend over **temporal intervals**, which are themselves **occurrents**. **Processes** also have **process boundaries**, their beginnings (and endings), which occupy **temporal instants**. Loving is best understood, as we will see, in terms of **processes** with proper process parts, occurring over a

---
[11] We adopt the convention of placing classes in bold and relations in italics.

**temporal interval** with proper parts, some of which are **temporal instants** that have **process boundary** occupants. That said, a full picture intimately involves the other BFO branch.

**Continuants** are entities that persist through time while maintaining their identity. They do not have temporal parts and may change their properties and other parts as they endure. **Independent continuants** are **continuants** which do not existentially depend on other entities for their existence [37].[12] **Material entity** – **independent continuants** which have matter as mereological parts - act as a parent class for **agent**, which is not strictly speaking part of BFO, but is reflected in several extensions of it. Agents, of course, play an important role in loving, and for our purposes can be understood as **material entities** that are able to engage in intentional actions [36].

**Independent continuants** are the bearers of properties, some of which fall under **specifically dependent continuant**, instances of which depend for their existence on instances of **independent continuant**. Such properties are "specific" in the sense that they may not migrate to other bearers. The smile on your face, for example, is a **specifically dependent continuant** that is uniquely yours. Indeed, it is a **quality** of yours, a subclass of **specifically dependent continuant**, instances of which - such as shapes and smiles - are fully manifested when manifested at all. There is nothing more to the shape of your smile than what is presented on your face with that shape. We posit that an **agent's** sensations are mental qualities which *inhere in* the **agent**. We also posit that the agent and their mental qualities are engaged in some manner of causal correlation to the object of love.[13]

**Qualities** stand in contrast to **dispositions**, which are **realizable entities**, a subclass of **specifically dependent continuants** that are not fully manifested when they exist, such as the irascibility of your neighbor [29]. If all goes well, you will never encounter this **disposition** manifesting; they are irascible, nonetheless. In BFO, we connect **dispositions** to the **occurrent** side of BFO by saying the **disposition** is *realized in* some **process**, such as being irascible. More relevant here is that an **agent** bearing sensations - understood as mental qualities – likely also bears some **disposition** to make judgments about those sensations. When an **agent** so disposed *realizes* this **disposition**, they *participate in* an act of judgment, which is a **process**. Though not immediately represented **CV4**, we might also characterize the inhibitory control underwriting the answerability for love as a **disposition** borne by relevant agents, realized in blocking the acquisition of certain **qualities**, other **dispositions**, or realization of other **processes**, analogous to the blocking disposition modeling of BFO in life science ontologies [40].

A sibling class of **independent continuant** and **specifically dependent continuant** is the class **generically dependent continuant**, instances of which are copyable patterns, existing only if *concretized in* some **continuant**. **Generically dependent** continuants are "generic" in the sense that they do not depend for their existence on any specific bearer, as they may be copied or transmitted across many. When an **agent** realizes a **disposition** to judge a sensation and so participate in an act of judgment, we understand the output of this judgment as a special type of **generically dependent continuant**, namely, an **information content entity**, which is a copyable pattern that *is about* some entity [36]. **Information content entities** are not strictly speaking part of BFO, but like **agents**, such entities are widely used in extensions. **Table 1**[14] highlights useful classes from the BFO ecosystem leveraged here.

---

[12]Ontological dependence, holding between *x* and *y* when the former cannot exist without the latter [38].
[13]What exactly this causal relationship amounts to must wait for another time; see [39] for discussion of the challenges associated with modeling causality and correlation in ontologies. Whatever the specification, for our purposes there should be the insistence on acquaintance with the beloved.

**Table 1**
Definitions/elucidations of selected ontology elements in the BFO ecosystem

| Elements | Elucidation/Definition |
|---|---|
| *Continuant* | An entity that persists, endures, or continues to exist through time while maintaining its identity. |
| *Independent Continuant* | A continuant which is such that there is no $x$ such that it specifically depends on $x$ and no $y$ such that it generically depends on $y$. |
| *Specifically Dependent Continuant* | A continuant which is such that **(i)** there is some independent continuant $x$ that is not a spatial region, and which **(ii)** specifically depends on $x$. |
| *Generically Dependent Continuant* | An entity that exists in virtue of the fact that there is at least one of what may be multiple copies. |
| *Material Entity* | An independent continuant that at all times at which it exists has some portion of matter as continuant part. |
| *Quality* | A specifically dependent continuant that, in contrast to roles and dispositions, does not require any further process in order to be realized. |
| *Realizable Entity* | A specifically dependent continuant that inheres in some independent continuant which is not a spatial region and is of a type some instances of which are realized in processes of a correlated type. |
| *Disposition* | A realizable entity such that if it ceases to exist, then its bearer is physically changed, and its realization occurs when and because this bearer is in some special physical circumstances, and this realization occurs in virtue of the bearer's physical make-up. |
| *Occurrent* | An entity that unfolds itself in time or is the start or end of such an entity or is a temporal or spatiotemporal region. |
| *Process* | An occurrent that has some temporal proper part and for some time has a material entity as participant. |
| *Temporal Instant* | A temporal instant is a zero-dimensional temporal region that has no proper temporal part |
| *Temporal Interval* | A temporal interval is a one-dimensional temporal region that is continuous, thus without gaps or breaks |
| *Agent* | A material entity that is able to engage in intentional acts. |
| *Information Content Entity* | A generically dependent continuant that generically depends on some information bearing entity and stands in relation of aboutness to some entity |

---

[14]Elucidations are descriptions provided to help fix the referent of primitive terms. Definitions express individually necessary and jointly sufficient conditions for an entity to be an instance of the class defined.

| *is about* | A primitive relationship between an information content entity and some entity |
|---|---|
| *x inheres in y* | x is a specifically dependent continuant & y is an independent continuant that is not a spatial region & x specifically depends on y |
| *participates in* | Holds between some x that is either a specifically dependent continuant or generically dependent continuant or independent continuant that is not a spatial region & some process y such that x participates in y some way |
| *x temporal part of y* | x occurrent part of y & (x and y are temporal regions) or (x and y are spatiotemporal regions & x temporally projects onto an occurrent part of the temporal region that y temporally projects onto) or (x and y are processes or process boundaries & x occupies a temporal region that is an occurrent part of the temporal region that y occupies) |

*3.2. Summary*

**CV4** can be ontologically anchored by directly mapping its components onto the terms and relations of BFO. According to **CV4**, love unfolds over time and includes sub-processes in which the lover experiences positive sensations causally correlated with the beloved with whom they have an actual acquaintance, judges both the sensations and the experiences in which they occur as valuable and exhibits inhibitory control over such judgments.

In BFO terms, the loving process unfolds over a **temporal interval** and is composed of **proper temporal parts** some of which are love events, which are **processes** mereologically bound to the larger loving process. States of loving at a time—such as being in love at an **instant** or over an **interval** smaller than the whole—can be understood as snapshots of the broader love **process**, that is, ontologically dependent on the extended process and deriving their status from it.

Prior to or perhaps simultaneous with love events, the loving **agent** *participates in* the loving process has *inhering in* them a positive sensation, a **mental quality**, that is *causally correlated* with the beloved. Having experienced such sensations, the lover obtains a **disposition** that, if realized, is *realized in* acts of judgment, which are themselves **processes**. These acts of judgment in turn generate **information content entities** that are about the beloved, reflecting assessment that the sensation experienced – and accordingly the person with whom the experience is causally connected – is valuable. Throughout, *inhering in* the **agent** is a **disposition** to inhibit or block "falling in love", whether that involve preventing new **sensations**, **dispositions**, or **processes**.

A love event, then, is a **process** involving an **agent** bearing a **mental quality** and a **disposition** to evaluate that **mental quality**. Each love event is a **temporal part** of the larger love process and contributes to what we called the loving sum, which here is the **process** that consists of all love events over the relevant **temporal interval**. Any mereological part of the interval during which such events do not occur, is what we called the loving complement. As discussed, the ratio of the loving sum to the entire interval must exceed some threshold *T* for the agent to count as loving the beloved during that interval. While *T* remains underspecified in our formula, we envision this being supplied by or operative with the act of judgment.

*3.3. Going Forward*

This ontological foundation, aligning **CV4** with BFO, is not without limitations. First, while BFO provides a rich set of categories for modeling temporal structure, participation, dependence, and realization, it does not natively capture certain normative features of love. Aspects such as mutuality and authenticity—important in ethical or interpersonal contexts—are not directly addressed within the current ontological vocabulary; a full picture of loving will undoubtably require integration with BFO extensions. This is clear when reflecting on the account of acquaintance which, while functionally modeled through *participates in* and causally linking **processes**, remains under formalized. Similar remarks apply to our underspecified use of causal correlation. Additionally, our proposal might be expanded to accommodate the role of lover, perhaps understood as a **realizable entity** that *inheres in* the **agent** and is *realized in* the loving process, as described by the appropriately named BFO class **role**.

There are other hallmarks of loving neither developed here nor anchored in an ontology, but worth incorporating in future work. For example, it is a necessary condition on loving such that if one loves someone else, then the lover believes the beloved should receive more goods, happiness, candy, and so on, than the beloved believes the beloved deserves. This feature goes some way to explain, among other things, why it is that I will sacrifice much more for my child than I would for my neighbor's child, despite each deserving respect as autonomous, dignified agents. For while I do not *believe* my child deserves more than the neighbor's child, I still *want* more for my child. It is here we find the irrationality of loving, which applies not only to family members but also romantic love. Future work explicating such principles will likely shed light on the normative aspects of love mentioned above.

Going forward, with the preceding observations accommodated, a formal axiomatization of **CV4** within OWL or first-order logic will also be needed, leveraging extensions in the BFO ecosystem where possible to promote interoperability and reuse. Empirical integration with affective science and social psychology will additionally be pursued, in the interest of ensuring the ontological model developed aligns with and applies to the research which motivates our reflections. This should include, moreover, cross-domain ontological consistency testing, to ensure that love as modeled here aligns with how related entities—emotion, commitment, normativity, and memory—are represented across ontology-driven systems.

**4. Conclusion**

This paper presents a formal ontological account of love, proposing that love is best understood as a temporally extended process involving both passive sensations and active evaluative judgments. We build on the Concatenation View (CV) ultimately arriving at a proposal which includes conditions of causal correlation and acquaintance to link the non-propositional nature of sensations with the content-bearing nature of judgment. Using Basic Formal Ontology (BFO), we model love as an occurrent composed of love events (temporal parts), qualities (sensations), realizable entities, acts of judgment, and information content entities (outputs of judgments), with thresholds over time determining whether love holds. The framework addresses both the passiveness and answerability of love, maintaining dispositions towards inhibitory control connect passive and active aspects. Given the breadth and scope of this topic, we can perhaps be forgiven for providing here a foundation for ontology engineering. It is our hope that these remarks spur greater interest in this topic; one cannot love alone, after all.